\documentclass[runningheads]{llncs}

\usepackage[T1]{fontenc}
\usepackage{graphicx}
\usepackage{forest}
\usepackage{xcolor}
\usepackage[utf8]{inputenc} 
\usepackage[T1]{fontenc}       
\usepackage{booktabs}       
\usepackage{amsfonts}       
\usepackage{nicefrac}     
\usepackage{microtype} 
\usepackage{lipsum}
\usepackage{fancyhdr}             
\graphicspath{{media/}}
\usepackage{array}
\usepackage{cite}
\usepackage{amsmath}

\usepackage{enumitem}
\usepackage{multirow}
\usepackage{multicol}
\usepackage{arydshln}
\usepackage{makecell}
\usepackage[normalem]{ulem}
\usepackage[hidelinks,breaklinks=true]{hyperref}
\usepackage{xurl}
\usepackage{float}
\usepackage{fancyvrb}
\usepackage{hyperref}
\hypersetup{
    colorlinks=true,
    linkcolor=blue,
    filecolor=blue,      
    urlcolor=blue,
    citecolor=blue,
}
\usepackage{fontawesome5}
\usepackage{subfig}

\newlist{rqs}{enumerate}{1}
\setlist[rqs]{label*=\textbf{RQ\arabic*}}

\begin{document}

\title{LeafData: An Agentic System for Data Migration}

\author{Sadanand Katukuri\inst{1}
\and
Rajasekhar Bada\inst{1}
\and
Navya Induri\inst{1}
\and
Rohit Gandham \inst{2}
\and
Lynette Pinto\inst{2}
\and
Joses Selvan\inst{3}
\and
Abishek Krishnamoorthy\inst{2}
\and
Joseph Rozario\inst{4}
\and
Pu Tian\inst{5}
\and
Pavan Poudel\inst{1}
\and
Yalong Wu\inst{1}\textsuperscript{*}}

\authorrunning{S. Katukuri et al.}

\institute{University of Houston--Clear Lake, Houston TX 77058, USA \\
\email{\{KatukuriS5734, BadaR4805, InduriN9196, Poudel, WuY\}@uhcl.edu}\\
\and
Infodat International Inc., Houston TX 77042, USA\\
\email{\{Rohit.Gandham, Lynette.Pinto, Abishek.Ravi\}@infodatinc.com}\\
\and
Cardinal Health, Dublin OH 43017, USA\\
\email{josessandeep.thamaraiselvan@cardinalhealth.com}\\
\and
Fontus Labs Inc., Houston TX 77042, USA\\
\email{joseph.rozario@morphusdata.com}\\
\and
Stockton University, Galloway NJ 08205, USA\\
\email{Pu.Tian@stockton.edu}}

\maketitle     
\begingroup\renewcommand\thefootnote{*}
\footnotetext{Corresponding author}

\begin{abstract}
Modern data migration relies on JSON configuration to define data connection, pipeline logic, and orchestration behavior. This requires domain knowledge from users and is time-consuming and error-prone. In this paper, we present LeafData, an agentic system that converts user intent into validated and executable JSON configuration for data migration. Specifically, LeafData comprises a frontend chatbot and the backend service. The chatbot  incrementally collects required information from users and performs schema-driven validation, while the backend service processes validated inputs and generates JSON configuration artifacts. These artifacts are directly consumable by orchestration platforms, enabling end-to-end pipeline generation and execution without manual coding. LeafData supports heterogeneous data migration across various data sources and connectors including relational databases, file-based systems, document-oriented databases, and REST APIs.

\end{abstract}

\section{Introduction}
\label{sec:introduction}

Data migration pipeline \cite{dehury2020ccodamic, morris2012practical} is used to define data connections, specify source and destination systems, configure schedules (e.g., one-time, daily, or cron-based intervals), and manage pipeline executions. These tasks are typically performed by orchestration platforms (e.g., Apache Airflow and Dagster) \cite{airflow2026coreconcepts, airflow2026dags, dagster2026sda, dagster2026assetsapi} using structured configuration formats such as JSON, web-based forms, or custom scripts \cite{airflow2026coreconcepts, airflow2026dags}. Specifically, a pipeline in orchestration platforms implements a directed acyclic graph (DAG) \cite{harenslak2021data, rodrigues2012graph} for an ordered set of dependent tasks. Each DAG node represents a task such as data extraction and transformation, while edges define execution dependencies between tasks. DAG enables reliable scheduling, execution monitoring, and failure handling (e.g., task retries and error tracking) in data migration pipeline. However, these platforms assume that users can provide complete and correct configurations, creating a barrier for non-expert users during the configuration stage.

To address this issue, we present LeafData\textsuperscript{\faGithub}, a chatbot-driven agentic system that converts natural language user intent into validated JSON configuration artifacts for data migration pipeline. Specifically, LeafData uses a frontend chatbot to incrementally collect information from the user and maintains interaction state over multiple turns. The user input is validated by Apache Avro using predefined schema that enforces required fields, data types, dependencies, etc. This reduces unnecessary user effort and ensures that all inputs are correct and satisfy constraints before configuration generation. Once validation is complete, the backend service generates JSON configuration artifacts for connection setup, pipeline (i.e., DAG) creation, and execution triggering. These artifacts are directly consumable by orchestration platforms such as Apache Airflow. 

\begingroup\renewcommand\thefootnote{\texttt{\faGithub}}
\footnotetext{\href{https://github.com/FontsLabs-Morphus/LeafData.git}{LeafData codebase with Morphus}}
\endgroup

LeafData supports data migration across multiple data sources and connectors, including relational databases (e.g., MySQL, Oracle, and MariaDB), file-based systems (e.g., AWS S3 Files, SFTP, and Excel), document-oriented databases (e.g., MongoDB), and REST APIs. Specifically, Leafdata processes both structured data (e.g., relational tables) and semi-structured data (e.g., JSON documents and API responses) within a unified pipeline framework. To enable this, LeafData constructs standardized JSON artifacts that define connection parameters, task dependencies, and execution behavior, allowing the orchestration layer to generate executable pipelines without requiring source-specific logic. Unlike existing approaches that focus on query generation or require complete user-provided configurations, LeafData combines guided chatbot interaction, schema-driven validation, and deterministic JSON configuration generation to improve usability, reduce configuration errors before execution, and ensure reproducibility of data migration pipelines. In other words, the same user inputs produce identical validated JSON configuration artifacts, which generate the same pipeline and execution results across runs.  

The remainder of this paper is organized as follows. In Section~\ref{sec:relatedwork}, we review related works. In Section~\ref{sec:systemmodel}, we present our system model. In Sections~\ref{sec:design} and~\ref{sec:service}, we describe frontend design and backend service, respectively. In Section~\ref{sec:payloads}, we explain supported JSON payloads. In Section~\ref{sec:casestudies}, we demonstrate our case study. Finally, we conclude our paper in Section~\ref{sec:conclusion}.

\section{Related Works}
\label{sec:relatedwork}

Existing research on natural language interfaces for data systems has focused on translating user inputs into executable queries, such as Text-to-SQL systems \cite{nli4db2025survey, text2sql2025survey}. These approaches enable users to interact with databases by mapping natural language user intent to structured query representations. This is effective for query generation but limited to single-query execution and does not address the broader problem of generating complete pipeline configurations, including connection setup, scheduling, and orchestration.

Workflow orchestration platforms such as Apache Airflow, Prefect, and Dagster provide robust frameworks to define and execute data pipelines using directed acyclic graphs (DAGs) \cite{airflow2026coreconcepts, airflow2026dags, dagster2026sda, dagster2026assetsapi}. These platforms support scheduling, execution monitoring, and failure handling for complex data workflows. However, they assume that users can create complete and correct JSON or code-based configurations. This introduces a barrier for non-expert users. Additionally, Great Expectations \cite{greatexpectations2024validation} enforces data quality constraints during pipeline execution to improve reliability but does not generate valid pipeline configurations from user intent. Similarly, emerging industry approaches integrate large language models into workflow orchestrations \cite{astronomer2024llmops, airflow2024mlops}. They focus on assisting execution and monitoring rather than producing validated configuration artifacts.

\section{System Model}
\label{sec:systemmodel}

\begin{figure*}[t]
\centering
\includegraphics[width=\textwidth]{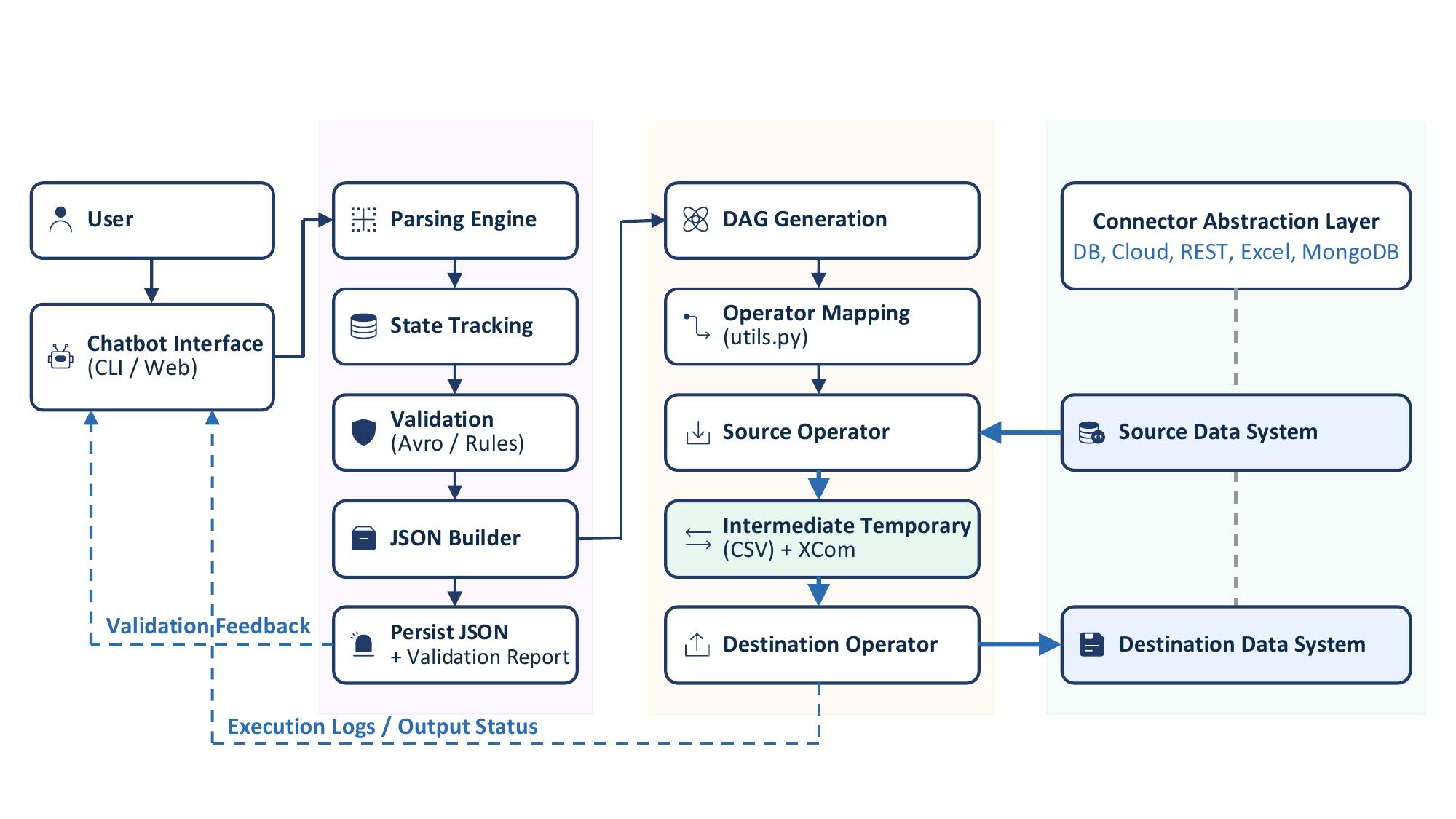}
\vspace{-11mm}
\caption{System Model of LeafData}
\vspace{-3mm}
\label{fig:system_architecture}
\end{figure*}

In this section, we present the overall architecture and execution flow of LeafData, as shown in Fig.~\ref{fig:system_architecture}. Specifically, LeafData follows an end-to-end workflow that begins with user interaction and extends to pipeline execution across heterogeneous data systems. The process is initiated through a chatbot that collects user intent and forwards it to the backend configuration service. 

The backend service processes user intent through parsing engine, state tracking, validation, and JSON builder to produce execution-ready configuration artifacts. These artifacts define connection parameters, pipeline (i.e., DAG) structure, and execution triggers that are subsequently used by Airflow orchestration to construct and execute data migration pipelines. Specifically, the backend service maintains a structured interaction state that represents configuration parameters as a set of fields categorized as known, missing, or invalid. User inputs are continuously incorporated into this internal representation, allowing LeafData to preserve prior validated values while updating relevant fields. This state-based approach ensures continuity across multiple interactions and avoids redundant data collection. Additionally, validation is enforced using predefined schema to ensure that all required fields satisfy type, format, and dependency requirements before configuration generation. LeafData  does not proceed to generate JSON artifacts unless all constraints are met. The JSON builder constructs execution-ready configuration artifacts that define connections, pipeline structure, and execution parameters. These artifacts are recorded along with validation metadata to enable traceability and reuse. Validation feedback generated during this process is communicated back to the user through the chatbot interface.

The generated JSON configuration artifacts are consumed by the Airflow orchestration to construct executable data pipeline. These artifacts provide structured definitions of pipeline metadata, task dependencies, source and destination configurations, and execution parameters. As shown in Fig.~\ref{fig:system_architecture}, the orchestration layer includes DAG generation, operator mapping, and execution operators, while the data and connectors layer comprises external date systems and a connector abstraction layer enabling interaction with heterogeneous data sources. The orchestration platform interprets the pipeline JSON to generate operator-level tasks, including source extraction, intermediate processing, and destination loading. For example, a pipeline JSON describes the data source and destination and how the data should be mapped. This information is used to create executable tasks that transfer data from the source to the destination.

Data transfer is performed through a staged pipeline. Specifically, data is extracted from the source system using a source operator, optionally processed through an intermediate stage, and loaded into the destination system using a destination operator. The connector abstraction layer provides a unified interface to interact with various data sources including relational databases, file-based systems, document-oriented databases, and REST-based services. This allows the orchestration layer to execute pipelines without source-specific logic. The orchestration layer also generates pipeline state, logs, and output status and communicates them back to the chatbot interface to enable pipeline execution monitoring and visibility without exposing underlying system complexity.

To sum up, LeafData is designed to bridge the gap between natural language interaction and pipeline execution. It provides a unified framework that connects user intent, configuration generation, and orchestration-based execution in a seamless and structured manner.

\section{Frontend Design}
\label{sec:design}

In this section, we describe LeafData's frontend model, focusing on how users interact with the system through the chatbot interface.

The chatbot interface enables users to initiate data migration using natural language. Specifically, a user submits a high-level request such as specifying source and destination data systems. The chatbot transitions into input collection, requests one field at a time, determines the next required input based on the current interaction state, and updates prompts accordingly. This gradually guides users from initial intent to complete configuration. In addition to standard input collection, the chatbot supports user-assisted control commands. For example, users can use inline queries \texttt{@ask} and \texttt{@change}  to request clarification during input collection and modify previously entered values, respectively. The interaction concludes with all collected inputs summarized before generating configuration artifacts.

Each user interaction is maintained within a session that persists throughout the conversation. The session stores user inputs, the current interaction state, and the next required field, enabling the chatbot to continue the interaction seamlessly across multiple turns. The session is updated as the user progresses. This allows the chatbot to track progress and resume from the correct step. Additionally, the chatbot generates prompts with concise format guidance, such as value patterns or examples, to ensure context-awareness and relevance to the current step of configuration. The chatbot can also provide targeted correction guidance and re-prompt for the same field whenever a user input does not satisfy required constraints. Any sensitive inputs such as credentials are masked.

\section{Backend Service}
\label{sec:service}

In this section, we present LeafData's backend design and focus on implementation level behavior including configuration processing, operator mapping, transformation logic, and integration with orchestration.

\subsection{Backend Configuration Service}\label{backend}

The backend configuration service, as shown in Fig.~\ref{fig:system_architecture}, is responsible for transforming user inputs into execution-ready pipeline configurations and coordinating their transition to the orchestration environment. It bridges the chatbot interface and the Airflow orchestration layer. The backend configuration service comprises a parsing engine, state tracking, a validation layer, and a JSON builder. These components collectively  interpret, verify, and convert user inputs into structured JSON confihuration artifacts that define data migration pipelines.

\textbf{Parsing Engine.}  
The backend begins with the parsing engine interpreting user input and extracting key configuration elements such as source system, destination system, and associated parameters. For example, a user request  \texttt{“transfer MariaDB to MySQL”} is parsed to \texttt{\{Source type: MariaDB\}} and \texttt{\{Destination type: MySQL\}}. This is implemented through connector extraction and normalization logic that map variations such as \texttt{“Maria DB”} or \texttt{“MySQL Server”} into standardized connector types.

\textbf{State Tracking.}  
The parsed and extracted values are stored in a state structure that represents the current configuration.  
For example, the state structure is as follows after parsing:
\begin{center}
\begin{BVerbatim}
{
  "source": { "type": "MariaDB" },
  "dest": { "type": "MySQL" },
  "mapping": {}
}.
\end{BVerbatim}
\end{center}
This state is updated incrementally as user provides additional inputs (e.g., host, port, and database). 

\textbf{Validation.}  
The validation layer enforces correctness constraints on user inputs before JSON configuration generation. It combines schema-based validation using Apache Avro with additional rule-based checks. Specifically, Avro schemas define the required structure, data types, and mandatory fields for each configuration component such as connection parameter or pipeline definition. Additionally, rule-based checks enforce constraints such as valid host formats, numeric port values, and consistency of dependency between fields. If any constraint is violated, the system immediately provides feedback through the chatbot interface and requests correction. This prevents the propagation of errors to downstream pipeline execution.

\textbf{JSON Builder.}  
Once all required inputs are collected and validated, the JSON builder constructs structured JSON artifacts including \texttt{source connection configuration}, \texttt{destination connection configuration}, \texttt{pipeline (i.e., DAG) definition}, and \texttt{execution trigger configuration}. For example, a \texttt{MariaDB-to-MySQL} transfer generates connection payloads with fields such as host, port, username, and database along with a DAG definition specifying the data transfer workflow.

\textbf{Persist JSON.}  
The generated JSON artifacts are saved to disk as part of the backend configuration service. These include \texttt{source\_connection.json}, \texttt{destination\_connection.json}, \texttt{create\_dag.json}, and \texttt{trigger\_dag.json}. The system masks sensitive values such as passwords during display but stores them in raw form in the generated files for execution purposes.
 
LeafData includes additional control logic within the chatbot to enable guided and reliable configuration. For example, the system requests one field at a time. If it detects a \texttt{MariaDB $\rightarrow$ MySQL transfer}, the system will prompt \texttt{Please enter Source (MariaDB) host:}. Each user input is validated immediately. If a user enters \texttt{Port: abc}, the system will respond
\texttt{Port must be numeric. Please try again.} Users can modify previously entered values without restarting the process. For example, \texttt{@change 1. Source (MariaDB) host} allows for updating the selected field. Users can also request clarification at any step without interrupting the workflow. For example, \texttt{@ask what is port?} provides a short explanation and resumes the current step.

\subsection{AirFlow Orchestration}\label{airflow}

The resulting configuration artifacts from Section~\ref{backend} are forwarded to Airflow orchestration, as shown in Fig.~\ref{fig:system_architecture}, where they are used for DAG generation and operator mapping. Based on these artifacts, executable workflows are constructed by instantiating source and destination operators and defining their dependencies. This enables  pipeline execution across heterogeneous data systems. 

\textbf{DAG Generation.}  
The system uses the generated configuration files such as  \texttt{create\_dag.json} to define a workflow in Apache Airflow. This corresponds to creating a DAG file that Airflow reads and displays in its interface. For example, if the configuration specifies a \texttt{MariaDB $\rightarrow$ MySQL transfer}, a DAG file named \texttt{mariadb\_to\_mysql} will be created with tasks representing extraction and loading steps. Once a user opens the Airflow user interface, this DAG will appear as a pipeline ready to be executed through a click on \texttt{Trigger DAG}.

\textbf{Operator Mapping.}  
Operator Mapping is responsible for assigning appropriate execution operators based on the connector types defined in the configuration. This mapping is implemented through a utility module \texttt{utils.py} that encapsulates connector-to-operator resolution logic. The utility layer abstracts the selection of operators by mapping source and destination types (e.g., MariaDB, SFTP, and REST API) to corresponding extraction and loading operators. This ensures that pipeline construction remains independent of connector-specific implementation details. For example, the mapping can include   
\texttt{MariaDB $\rightarrow$ Source database operator}, \texttt{SFTP $\rightarrow$ File extraction operator}, \texttt{REST API $\rightarrow$ API request operator}, and \texttt{MySQL $\rightarrow$ Destination loading operator}. This abstraction enables extensibility, allowing new connectors to be supported by updating the mapping logic without modifying the pipeline definition structure.

\textbf{Source Operator.}  
The first task in the DAG connects to the source system and extracts data. For example, the MariaDB operator runs a query such as \texttt{SELECT * FROM source\_table;} and retrieves rows from the database.

\textbf{Intermediate Temporary.}  
The extracted data is written to a temporary CSV file and the file path is passed between tasks using Airflow’s XCom. For example, the source task writes \texttt{/tmp/data.csv}, pushes it via XCom, and the destination task retrieves it. This avoids directly passing large datasets between tasks and ensures scalability.

\textbf{Destination Operator.}  
The destination operator reads the intermediate CSV file and loads data into the target system. For example, the MySQL operator reads \texttt{/tmp/data.csv} and inserts rows into \texttt{target\_table}.

Airflow tracks task status and logs during execution. Specifically, each task shows success/failure status, execution time, and logs for debugging. These logs can optionally be integrated back into the chatbot interface for user visibility. 

\subsection{Connector Abstraction and Data Sources}

As shown in Fig.~\ref{fig:system_architecture}, LeafData interacts with external data systems through a connector abstraction layer that acts as an interface between the orchestration pipeline and heterogeneous data sources, including relational databases (e.g., MariaDB, MySQL, and PostgreSQL), file-based systems (e.g., CSV, Excel, and SFTP), and REST-based services (e.g., API endpoints). This layer sits above the source and destination data systems and isolates the orchestration from the underlying implementation details of each system, allowing pipelines to operate uniformly regardless of data source types. In the implementation, each connector type is handled by an operator that encapsulates connection setup, data extraction, and data loading logic.

For example, pipelines \texttt{MariaDB $\rightarrow$ MySQL} and \texttt{SFTP $\rightarrow$ MySQL} have different data sources, but they will be executed using the same workflow: a source operator retrieves data, data is passed through an intermediate stage (CSV + XCom), and destination operator loads data into MySQL. The only difference lies in the connector logic. In \texttt{MariaDB $\rightarrow$ MySQL}, the source operator connects to a database and executes SQL queries. In \texttt{SFTP $\rightarrow$ MySQL}, the source operator connects to an SFTP server and reads files. This design allows the system to construct pipelines without embedding source-specific logic in the workflow. Thus, new data sources can be integrated by implementing additional operators within the connector abstraction layer, without modifying the overall pipeline structure.

\begin{figure*}[t]
\centering
\includegraphics[width=1\linewidth]{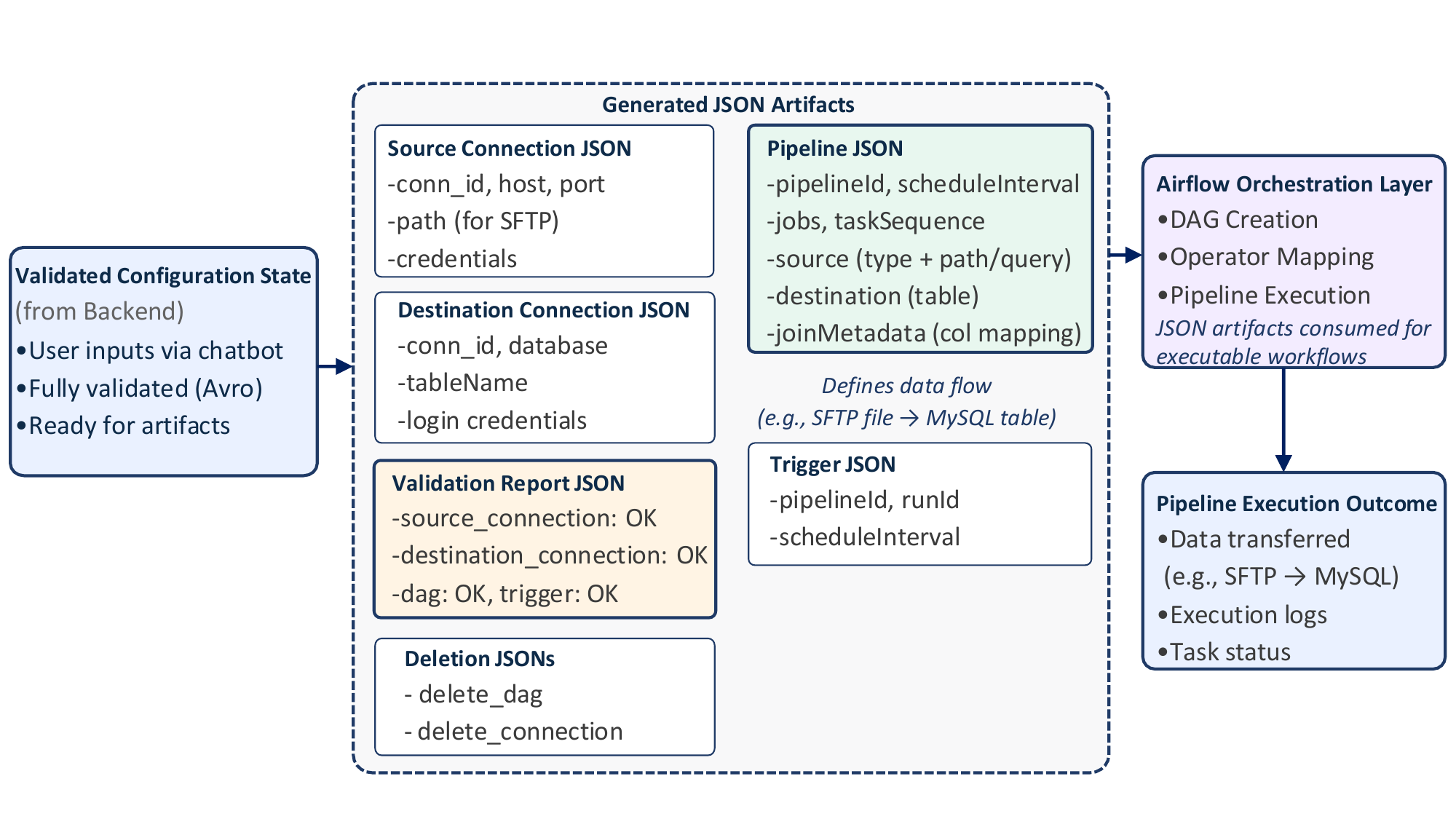}
\vspace{-0.7cm}
\caption{Structured JSON Payloads}
\label{fig:jsonflow}
\end{figure*}

\section{Supported JSON Payloads}\label{sec:payloads}

This section describes the structured JSON artifacts generated by LeafData, as illustrated in Fig.~\ref{fig:jsonflow}. These artifacts are produced by the backend configuration service and serve as the inputs to the AirFlow orchestration layer for pipeline creation and execution. In the implementation, each artifact is generated as a separate JSON file (e.g., \texttt{source\_connection.json}) and represents a specific configuration responsibility within the pipeline lifecycle.

\textbf{Connection Payloads.}
These payloads define the connection between LeafData and the external data systems and are generated as \texttt{source\_connection.json} and \texttt{destination\_connection.json}. Each file contains parameters such as host, port, credentials, and resource identifiers required to establish connections. For an example of an \texttt{SFTP $\rightarrow$ MySQL} pipeline, \texttt{source\_connection.json} stores SFTP host (e.g., \texttt{sftp.example.com}) and file path (e.g., \texttt{/home/foo/upload/books.csv}), while \texttt{destination\_connection.json} stores MySQL host, database name, and target table (e.g., \texttt{books\_demo\_copy}). These files are used by the orchestration layer to establish connections automatically during execution.

\textbf{Pipeline Payload.}
The pipeline configuration is generated as \texttt{create\_dag.json}, which defines how data flows from source to destination. This file contains pipeline identifier, scheduling information, task definitions, column mappings (e.g., \texttt{joinMetadata}), and execution order (e.g., \texttt{taskSequence}). For example, an \texttt{SFTP $\rightarrow$ MySQL transfer} has source: CSV file at \texttt{/home/foo/upload/books.csv}, destination: MySQL table \texttt{books\_demo\_copy}, and mapping: \texttt{col1 $\rightarrow$ colA}. This JSON file is read by the orchestration layer to generate a DAG in Airflow. Each entry in \texttt{taskSequence} becomes a task (e.g., extract or load) to define the execution workflow.

\textbf{Trigger Payload.}
The execution trigger is generated as \texttt{trigger\_dag.json}, which is responsible for starting the pipeline. This file includes pipeline ID, run ID, and scheduling interval. For example, if the pipeline is scheduled as daily, this file will trigger the DAG execution in Airflow either manually everyday or based on the defined schedule.

\textbf{Deletion Payloads.} In LeafData,
deletion operations are supported through JSON payloads such as \texttt{delete\_dag.json} and \texttt{delete\_connection.json}. These files contain identifiers required to remove existing configurations. For example, if a pipeline needs to be updated, \texttt{delete\_dag.json} will remove the existing DAG before a new one is created.

\textbf{Validation Report.}
The system also generates a validation summary file (i.e., \texttt{validation\_report.json}) that verifies all artifacts before execution. This file indicates the status of source connection, destination connection, pipeline definition, and trigger configuration. For example,
\begin{center}
\begin{BVerbatim}
{
  "source_connection": "OK",
  "destination_connection": "OK",
  "dag": "OK",
  "trigger": "OK"
}.
\end{BVerbatim}
\end{center}
This ensures that only valid configurations are passed to the orchestration layer, preventing runtime failures.

As shown in Fig.~\ref{fig:jsonflow}, the generated JSON files are consumed by the orchestration layer to construct and execute pipelines. Specifically, the \texttt{create\_dag.json} is used to generate DAG in Airflow. The \texttt{source\_connection.json} and \\ \texttt{destination\_connection.json} are used by operators to connect to data systems. The \texttt{trigger\_dag.json} initiates execution. Through this file-driven mechanism, LeafData converts user intent into structured JSON artifacts that directly drive pipeline creation and execution, ensuring consistency and reproducibility.

\begin{figure}[t]
\centering
\makebox[\textwidth][c]{\includegraphics[width=1.9\linewidth]{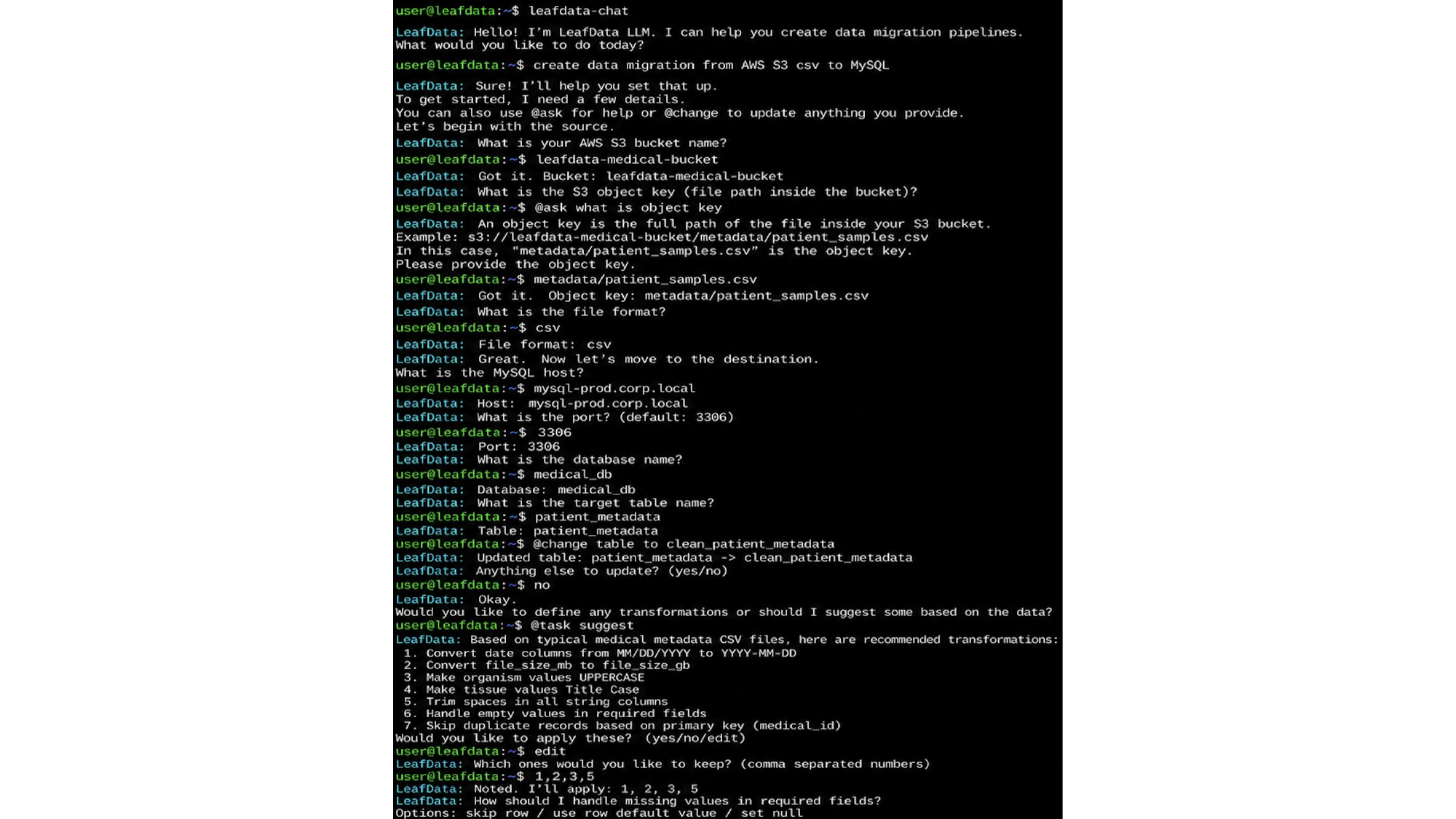}}
\vspace{-0.2cm}
\caption{Conversational Pipeline Configuration Part I}
\vspace{-0.5cm}
\label{fig:conversation 1}
\end{figure}

\begin{figure}[t]
\centering
\makebox[\textwidth][c]{\includegraphics[width=1.9\linewidth]{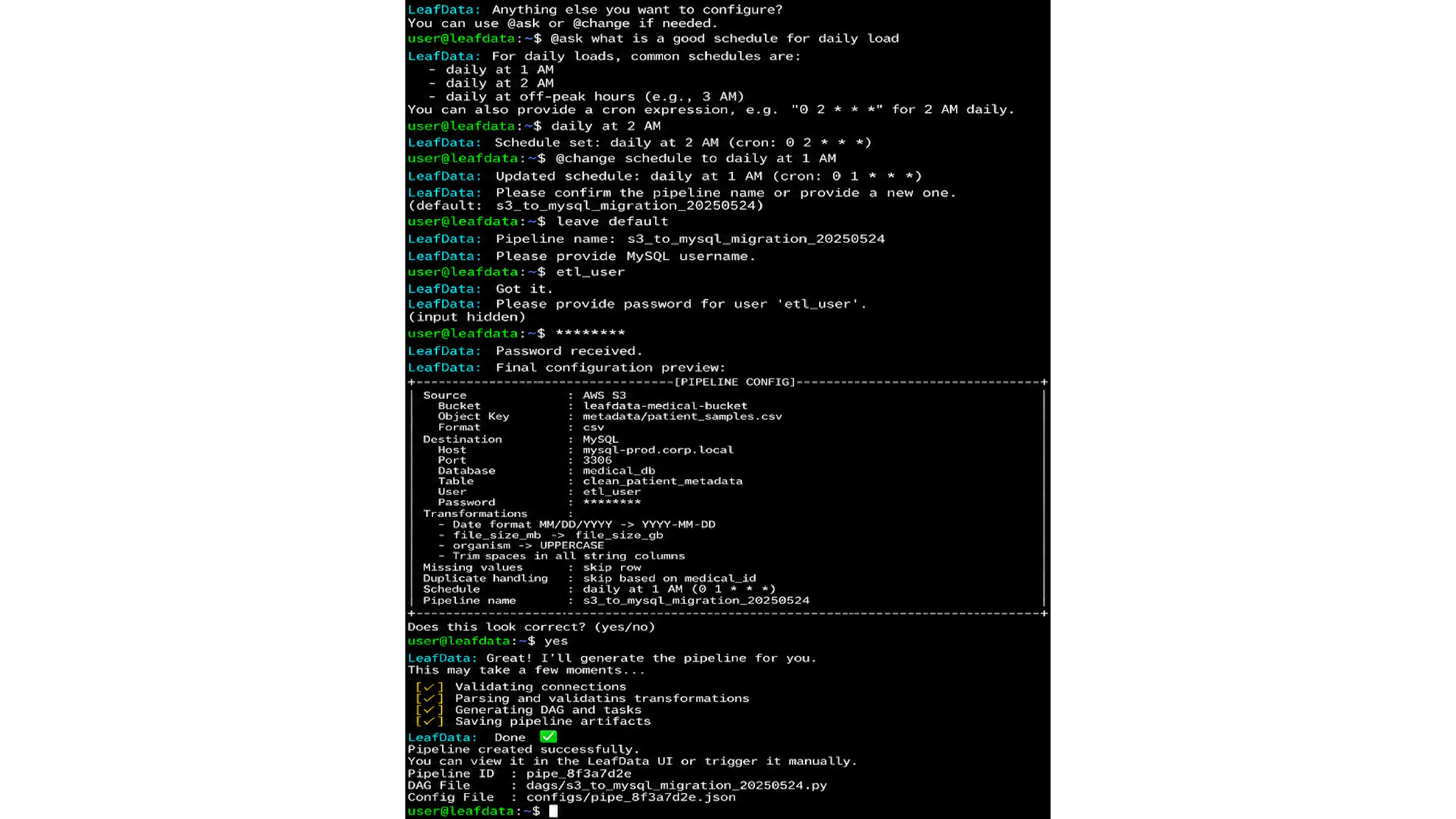}}
\vspace{-0.2cm}
\caption{Conversational Pipeline Configuration Part II}
\vspace{-0.5cm}
\label{fig:conversation 2}
\end{figure}

\begin{figure*}[t]
\centering
\makebox[\textwidth][c]{
\includegraphics[width=1.2\linewidth]{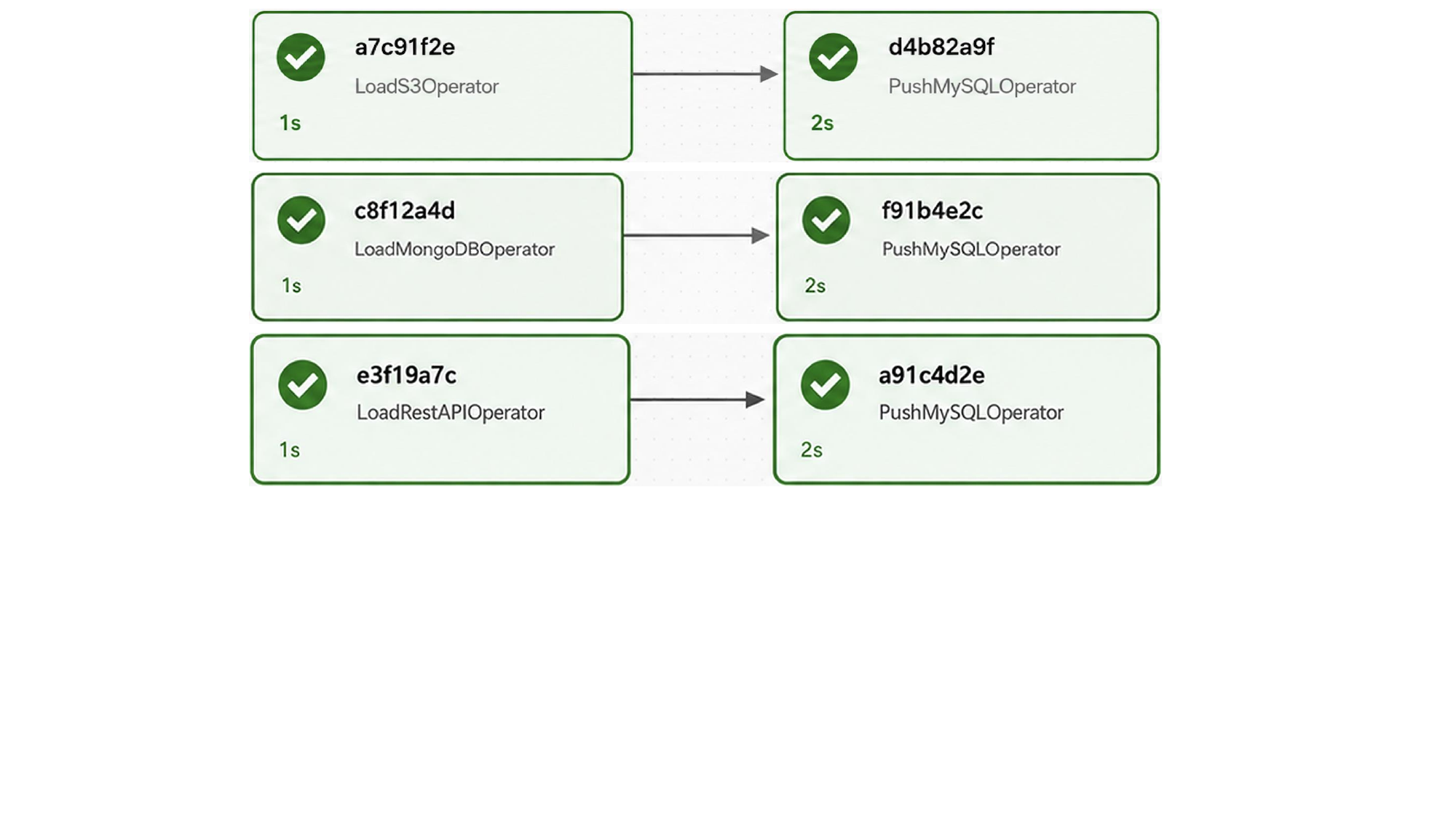}}
\vspace{-3.5cm}
\caption{Airflow Execution Graphs}
\vspace{-0.1cm}
\label{fig:executiongraphs}
\end{figure*}

\begin{figure}[t]
\centering
\makebox[\textwidth][c]{\includegraphics[width=1.3\linewidth]{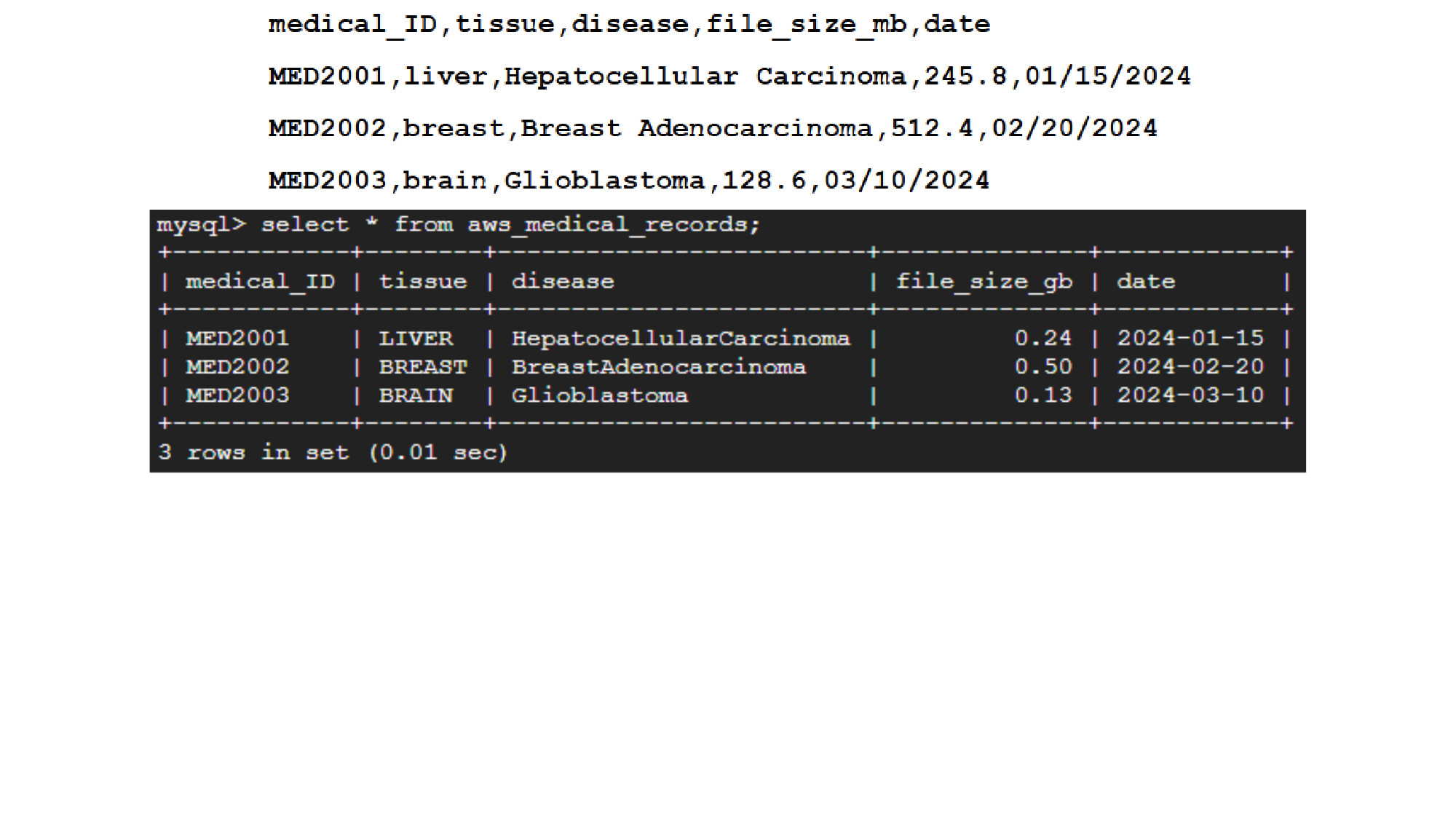}}
\vspace{-3.8cm}
\caption{AWS S3 to MySQL Migration}
\vspace{-0.5cm}
\label{fig:aws}
\end{figure}

\section{Case Study}\label{sec:casestudies}

In this section, we evaluate LeafData using a case study on data migration across various data sources and connectors. The goal is to demonstrate how user intent is translated into executable data pipelines. GPT 5.4 mini was used in our case.

As shown in Figs.~\ref{fig:conversation 1} and ~\ref{fig:conversation 2} for an AWS S3 to MySQL migration, LeafData begins with using a chatbot to collect required information such as bucket name, object key, database name, transformation rules, and execution schedule. The selected transformations comprise date format conversion, file size unit conversion, capitalization, and space trimming. LeafData additionally supports user-assisted interaction commands (i.e., \texttt{@ask} and \texttt{@change}) during configuration. Specifically, the @ask command allows users to request clarification without interrupting the workflow, and a user may enter \texttt{@ask what is object key} to request a short explanation. The \texttt{@change} command allows users to modify previously entered values. For example, \texttt{@change table to clean\_patient\_metadata} updates the stored value from \texttt{patient\_metadata} to \texttt{clean\_patient\_metadata}. LeafData eventually uses the collected information to generate backend configuration artifacts including source connections, destination connections, DAG definitions, and execution triggers, and the workflow proceeds to data pipeline orchestration.

The upper part of Fig.~\ref{fig:executiongraphs} shows the generated AWS S3-to-MySQL graph in Airflow. Specifically, LeafData uses the first task \texttt{LoadS3Operator} to retrieve file from AWS S3 bucket, stages it as an intermediate file, and uses the second task \texttt{PushMySQLOperator} to load the file into MySQL. The directed edge between the two tasks indicates that source extraction must complete before destination loading begins. The green task boxes and check markers signal a successful execution. Fig.~\ref{fig:aws} shows the actual file-based ingestion from the AWS S3 bucket into a MySQL table. The bucket contains a CSV file that has five attributes \texttt{medical\_ID}, \texttt{tissue}, \texttt{disease}, \texttt{file\_size\_mb}, and \texttt{date} for each medical record, as shown in the upper part of Fig.~\ref{fig:aws}. The \texttt{LoadS3Operator} reads the file, while the \texttt{PushMySQLOperator} loads it into the destination table after data transformation, as shown in the lower part of Fig.~\ref{fig:aws}.

To be comprehensive, we likewise examined data migration from MongoDB and REST API to MySQL, respectively. Fig.~\ref{fig:mongo} highlights data migration between document-based and relational databases. The middle part of Fig.~\ref{fig:executiongraphs} shows the generated MongoDB-to-MySQL graph in Airflow. MongoDB documents contain nested structures and system fields, as shown in the upper part of Fig.~\ref{fig:mongo}. LeafData automatically removes non-relational fields and flattens the structure before loading them into MySQL, as shown in the lower part of Fig.~\ref{fig:mongo}.

\begin{figure}[t]
\centering
\makebox[\textwidth][c]{\includegraphics[width=1.3\linewidth]{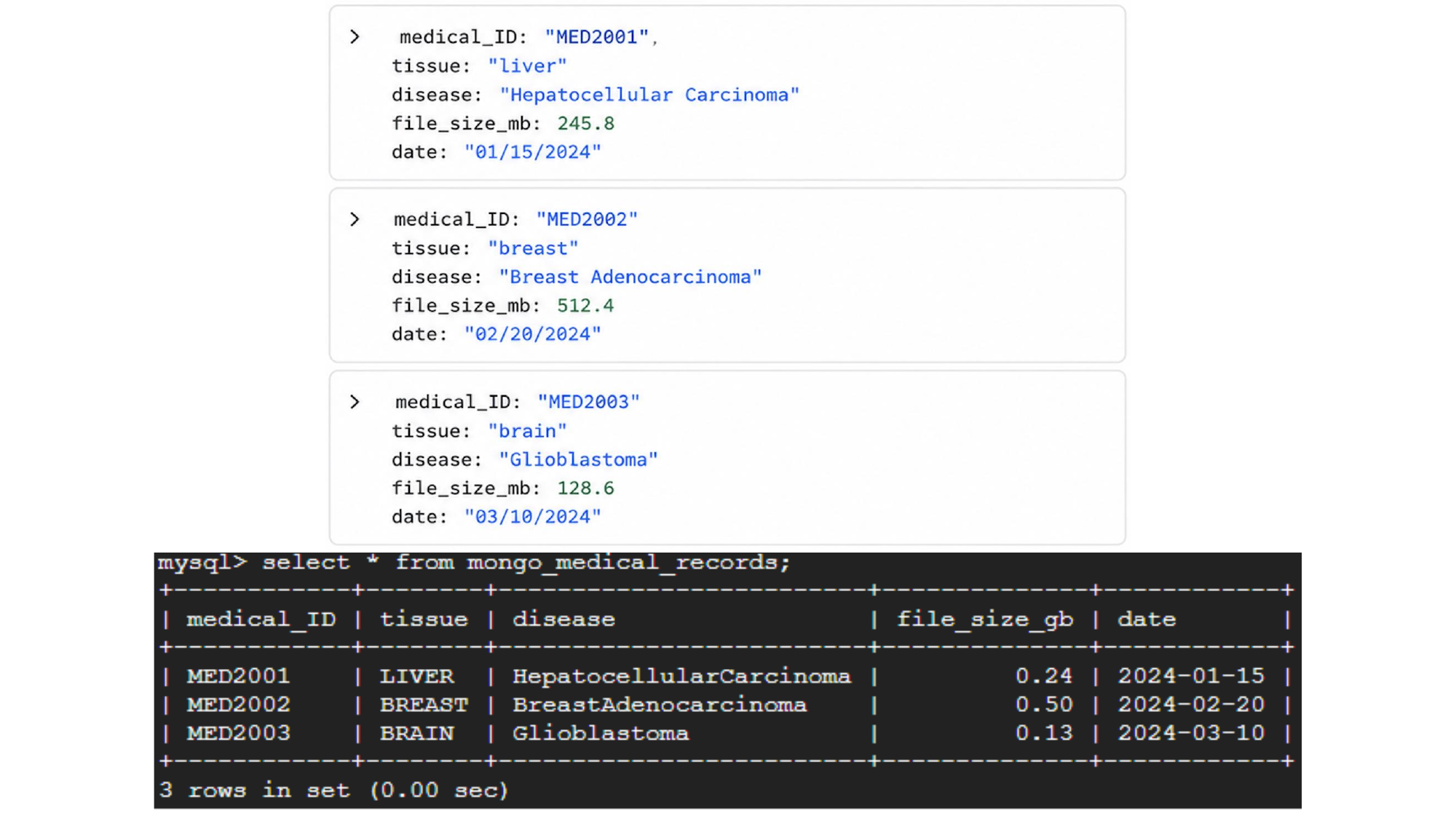}}
\vspace{-0.2cm}
\caption{MongoDB to MySQL Migration}
\vspace{-0.9cm}
\label{fig:mongo}
\end{figure}

Fig.~\ref{fig:restapi} demonstrates ingestion from a REST API into a MySQL table. The source API returns semi-structured JSON records, while the destination expects structured rows with fixed columns. Specifically, a user issues a request \texttt{``transfer medical record API data to MySQL''}, and LeafData collects the REST API endpoint, MySQL connection detail, target table name, and execution schedule through the chatbot. The lower part of Fig.~\ref{fig:executiongraphs} shows the generated REST API-to-MySQL graph in Airflow. As shown in the upper part of Fig.~\ref{fig:restapi}, the source REST API returns a JSON array containing medical records. LeafData stores the REST API endpoint in the generated pipeline configuration and maps the source connector to \texttt{LoadRestAPIOperator}. This operator sends a request to the API endpoint during execution, parses the JSON response, normalizes the records into tabular form, and stages the result as an intermediate CSV file. The staged file path is passed to the destination task through Airflow's XCom, and the destination task \texttt{PushMySQLOperator} reads the staged file and inserts it into the target MySQL table. The lower part of Fig.~\ref{fig:restapi} confirms the final result where the semi-structured API records are stored as relational rows in MySQL. 

\begin{figure}[t]
\centering
\makebox[\textwidth][c]{\includegraphics[width=1.3\linewidth]{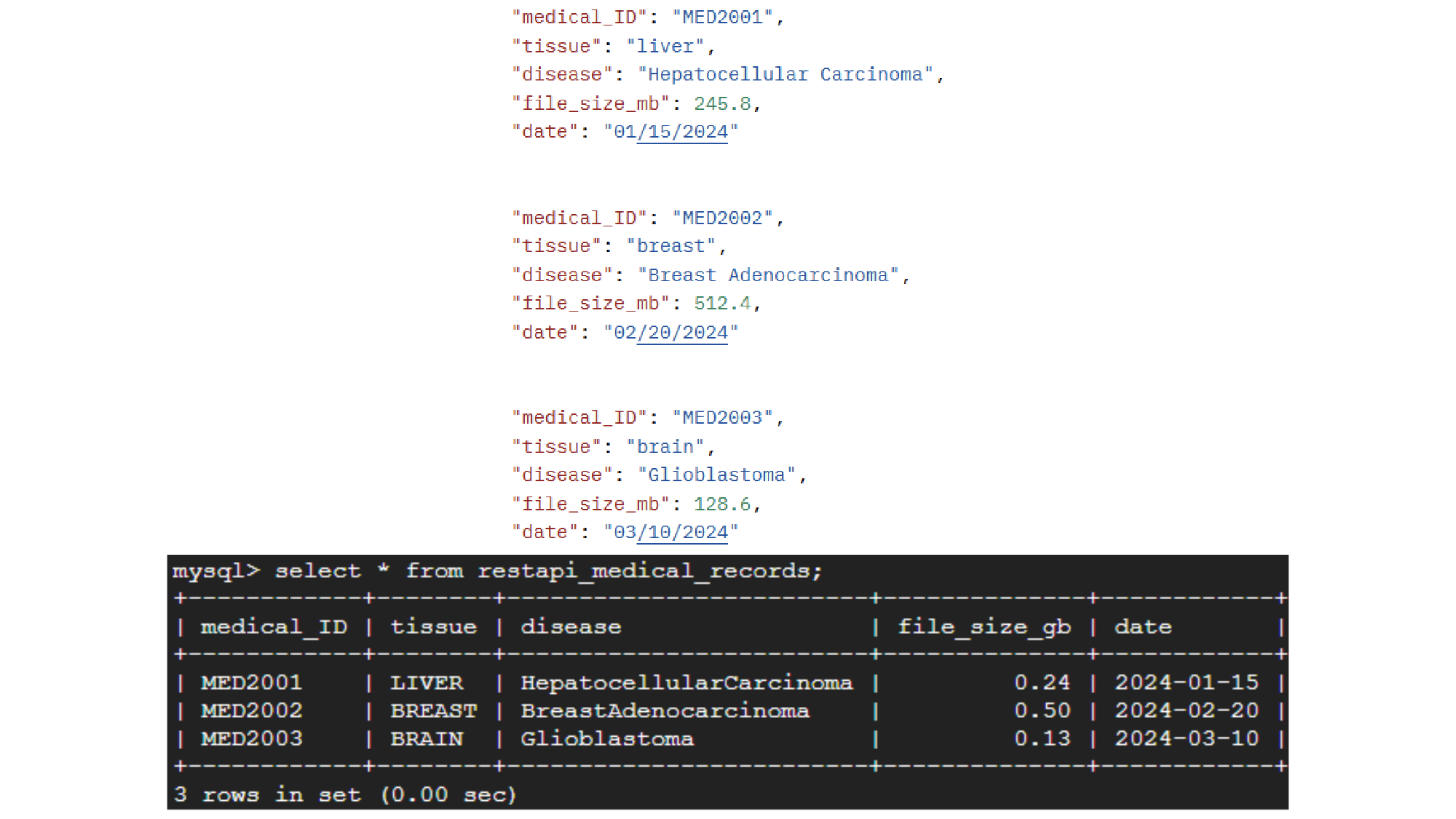}}
\vspace{-0.2cm}
\caption{REST API to MySQL Migration}
\vspace{-0.3cm}
\label{fig:restapi}
\end{figure}

\section{Conclusion}\label{sec:conclusion}

This paper presented LeafData, an agentic system that transforms natural language user inputs into validated JSON configuration artifacts for data migration pipelines. LeafData supports heterogeneous data sources and connectors and integrates with orchestration platform to generate and execute pipelines. This approach improves usability by reducing configuration effort, enhances reliability through early validation, and ensures reproducibility of pipeline execution. It is worth noting that LeafData supports primarily linear pipelines with single source and destination. Future work will focus on more complex workflows involving branching, conditional logic, and multi-stage dependencies.

\bibliographystyle{splncs04}
\bibliography{ref}

@misc{airflow2024mlops,
  author       = {{RNHTTR}},
  title        = {Real-world MLOps with Airflow 3.0 and the Airflow AI SDK},
  howpublished = {\url{https://medium.com/apache-airflow/real-world-mlops-with-airflow-3-0-and-the-airflow-ai-sdk-8c4a236e4d6b}},
  year         = {2025},
  note         = {Accessed: 2026-05-25}
}

@misc{airflow2026coreconcepts,
  author       = {{Apache Airflow}},
  title        = {Core Concepts},
  howpublished = {\url{https://airflow.apache.org/docs/apache-airflow/stable/core-concepts/index.html}},
  year         = {2026},
  note         = {Accessed: 2026-05-25}
}

@misc{airflow2026dags,
  author       = {{Apache Airflow}},
  title        = {Dags},
  howpublished = {\url{https://airflow.apache.org/docs/apache-airflow/stable/core-concepts/dags.html}},
  year         = {2026},
  note         = {Accessed: 2026-05-25}
}

@misc{astronomer2024llmops,
  author       = {{Astronomer}},
  title        = {DAY-2 OPERATIONS FOR LLM APPLICATIONS WITH APACHE AIRFLOW},
  howpublished = {\url{https://www.astronomer.io/blog/day-2-operations-for-llms-with-apache-airflow/}},
  year         = {2023},
  note         = {Accessed: 2026-05-25}
}

@misc{dagster2026assetsapi,
  author       = {{Dagster}},
  title        = {Assets},
  howpublished = {\url{https://docs.dagster.io/api/dagster/assets}},
  year         = {2026},
  note         = {Accessed: 2026-05-25}
}

@misc{dagster2026sda,
  author       = {{Dagster}},
  title        = {Software-defined Asset},
  howpublished = {\url{https://dagster.io/glossary/software-defined-assets}},
  year         = {2026},
  note         = {Accessed: 2026-05-25}
}

@article{dehury2020ccodamic,
  title={Ccodamic: A framework for coherent coordination of data migration and computation platforms},
  author={Dehury, Chinmaya Kumar and Srirama, Satish Narayana and Chhetri, Tek Raj},
  journal={Future Generation Computer Systems},
  volume={109},
  pages={1--16},
  year={2020},
  publisher={Elsevier}
}

@misc{greatexpectations2024validation,
  author       = {{Great Expectations}},
  title        = {Validate Data},
  howpublished = {\url{https://docs.greatexpectations.io/docs/0.18/oss/guides/validation/validate_data_lp/}},
  year         = {2024},
  note         = {Accessed: 2026-05-25}
}

@book{harenslak2021data,
  title={Data pipelines with apache airflow},
  author={Harenslak, Bas P and De Ruiter, Julian},
  year={2021},
  publisher={Simon and Schuster}
}

@book{morris2012practical,
  title={Practical data migration},
  author={Morris, Johny},
  year={2012},
  publisher={BCS, The Chartered Institute}
}

@article{rodrigues2012graph,
  title={A graph-based approach for designing extensible pipelines},
  author={Rodrigues, Ma{\'\i}ra R and Magalh{\~a}es, Wagner CS and Machado, Moara and Tarazona-Santos, Eduardo},
  journal={BMC bioinformatics},
  volume={13},
  number={1},
  pages={163},
  year={2012},
  publisher={Springer}
}

@article{nli4db2025survey,
  title={Nli4db: A systematic review of natural language interfaces for databases},
  author={Liu, Mengyi and Xu, Jianqiu},
  journal={arXiv preprint arXiv:2503.02435},
  year={2025}
}

@article{text2sql2025survey,
  title={A Survey of Text-to-SQL in the Era of LLMs: Where are we, and where are we going?},
  author={Liu, Xinyu and Shen, Shuyu and Li, Boyan and Ma, Peixian and Jiang, Runzhi and Zhang, Yuxin and Fan, Ju and Li, Guoliang and Tang, Nan and Luo, Yuyu},
  journal={IEEE Transactions on Knowledge and Data Engineering},
  year={2025},
  publisher={IEEE}
}
\end{document}